\pdfoutput=1

\documentclass[11pt]{article}

\usepackage[final]{acl}

\usepackage{times}
\usepackage{latexsym}
\usepackage{algorithm}
\usepackage{algorithmic}
\usepackage[T1]{fontenc}

\usepackage[utf8]{inputenc}

\usepackage{microtype}

\usepackage{inconsolata}

\usepackage{graphicx}

\usepackage{amsmath}
\usepackage{amsfonts}
\usepackage{multirow, makecell, booktabs}
\usepackage{listings}
\usepackage{xcolor}

\lstdefinestyle{PythonStyle}{
    language=Python,
    backgroundcolor=\color{lightgray!10}, 
    commentstyle=\color{green!50!black}\ttfamily, 
    keywordstyle=\color{blue}\bfseries, 
    stringstyle=\color{red}\ttfamily, 
    basicstyle=\ttfamily\footnotesize, 
    numberstyle=\tiny\color{gray}, 
    stepnumber=1, 
    showspaces=false, 
    showstringspaces=false, 
    showtabs=false, 
    frame=single, 
    rulesepcolor=\color{gray}, 
    breaklines=true, 
    captionpos=b, 
    tabsize=4, 
}
\title{Efficient Beam Search for LLMs Using Trie-Based Decoding}

\author{
Brian J Chan\textsuperscript{1$\dagger$} \quad
MaoXun Huang\textsuperscript{2$\dagger$} \quad
Jui-Hung Cheng\textsuperscript{1} \quad
Chao-Ting Chen\textsuperscript{1} \AND
Hen-Hsen Huang\textsuperscript{3} \\
\textsuperscript{1}Department of Computer Science, National Chengchi University, Taiwan\\
\textsuperscript{2}Department of Computer Science, Cornell University, U.S.\\
\textsuperscript{3}Institute of Information Science, Academia Sinica, Taiwan\\
\texttt{110703065@g.nccu.edu.tw} \quad
\texttt{mh2653@cornell.edu} \\
\texttt{\{110703007,110703038\}@g.nccu.edu.tw} \quad 
\texttt{hhhuang@iis.sinica.edu.tw} \\
\textsuperscript{$\dagger$}Equal contribution\\
}

\begin{document}
\maketitle
\begin{abstract}
This work presents a novel trie (prefix-tree)-based parallel decoding method that addresses the memory inefficiency of batch-based beam search. 
By sharing a single KV cache across beams with common prefixes, our approach dramatically reduces memory usage and enables efficient decoding. 
We evaluated our method across three attention architectures, Multi-Head Attention (Phi-3.5-mini-instruct), Grouped Query Attention (Llama-3.1-8B-Instruct), and Sliding Window Attention (Mistral-Small-24B-Instruct-2501), using CNN/DailyMail for abstractive summarization and HumanEval for code generation. 
Our experiments demonstrate substantial memory savings (4--8$\times$) and up to 2.4$\times$ faster decoding, without compromising generation quality. 
These results highlight our method's suitability for memory-constrained environments and large-scale deployments.
\end{abstract}
\section{Introduction}
Large language models (LLMs) face significant deployment challenges due to their high memory requirements. 
For example, the 8-billion-parameter Llama 3.1 model, when deployed in float16 precision, requires approximately 15.7GB of GPU memory solely for its model parameters. 
Processing an 8k token sequence adds another 2.5GB for the key-value (KV) cache. These constraints make efficient memory utilization a critical factor in optimizing LLM performance.

Memory optimization not only reduces hardware requirements but also accelerates inference. Modern GPUs often exhibit faster computation speeds than memory transfer rates, leading to a memory-bound performance bottleneck. 
Addressing this bottleneck has spurred innovations like Flash Attention~\citep{dao2022flashattention,dao2023flashattention2}, which minimizes memory operations. Efficient memory usage reduces the overhead of transferring data within the GPU, enhancing both speed and scalability.

The decoding process plays a pivotal role in the performance and quality of sequence generation in LLMs. Typical decoding strategies fall into three categories: greedy decoding, top-$k$ sampling, and beam search. 
Greedy decoding selects the most probable token at each step, offering simplicity but often failing to recover from suboptimal decisions. Top-$k$ sampling introduces diversity by choosing the next token from the $k$ most probable options based on their probabilities. 
While effective for generating varied outputs, top-$k$ sampling is prone to hallucination~\citep{dziri-etal-2021-neural}, limiting its applicability for tasks requiring high factual accuracy, such as programming, math, or retrieval-augmented generation (RAG)~\citep{lewis2021retrievalaugmentedgenerationknowledgeintensivenlp,pham2024reliablemedicalquestionanswering}.

Beam search, on the other hand,  maintains multiple candidate sequences (beams) at each time step and ultimately selects the one with the highest overall probability. Unlike greedy search, beam search can ``look ahead'' to identify sequences that may start with lower-probability tokens but lead to better overall outcomes. By keeping multiple hypotheses, beam search can recover from locally suboptimal decisions, often yielding better results than greedy decoding in certain tasks that require high accuracy, like recommendation~\citep{li2023gpt4recgenerativeframeworkpersonalized} and coding. 
However, its computational cost and memory demands make it less practical for real-world applications, especially at scale.

Beam search's high memory consumption stems from its handling of KV caches. 
While beam search explores multiple branches originating from a shared prefix, conventional batch-based implementations allocate independent KV caches for each branch, leading to significant memory redundancy, as overlapping tokens across branches are stored multiple times. 
Such inefficiencies make memory optimization crucial for scalable and cost-effective deployment.

\begin{figure}[!thb]
    \centering
    \includegraphics[width=1\linewidth]{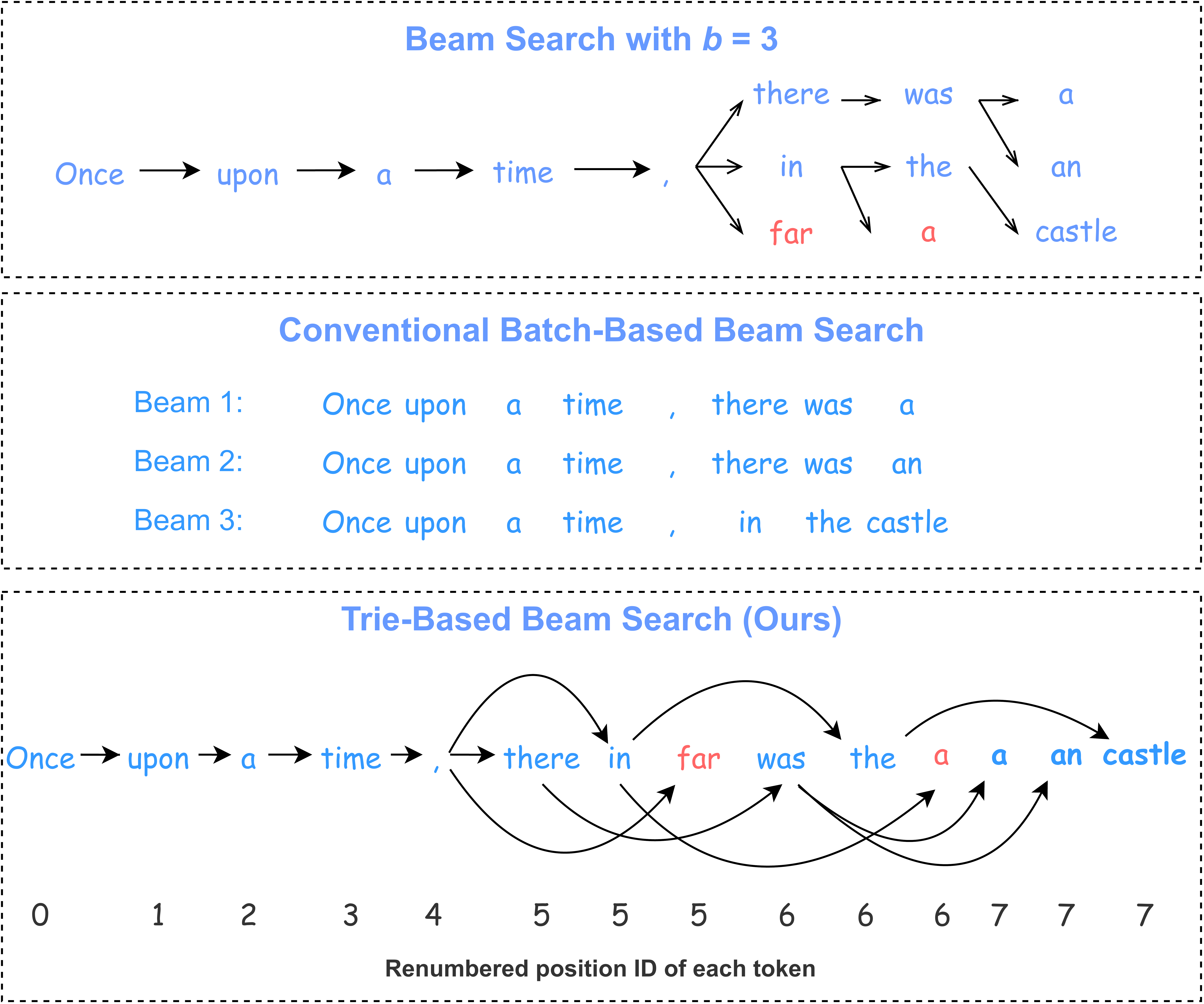}
    \caption{With beam width $b=3$, the top panel shows multiple explored hypotheses. 
    The middle illustrates conventional beam search, which stores redundant prefixes (e.g., ``Once upon a time,'') in separate caches. 
    The bottom shows our trie-based approach, consolidating overlaps into a shared trie to reduce memory while preserving beam width and behavior. 
    Red tokens (e.g., ``far'', ``a'') indicate pruned branches, and position IDs are reassigned to match conventional beam search.}
    \label{fig:bs-comparison}
\end{figure}

In this paper, we propose a novel trie (prefix tree)-based decoding method that significantly reduces memory usage in beam search by leveraging the hierarchical structure of shared prefixes among branches. 
Our approach consolidates all branches into a single shared kernel using a trie search strategy. 
As shown in Figure~\ref{fig:bs-comparison}, this method stores only unique tokens corresponding to shared prefixes, reducing memory consumption by eliminating redundant KV cache entries. 
For instance, in the illustrated example, our approach requires storing only 12 tokens, compared to the 21 tokens required by conventional batch-based methods.

The idea of trie-based decoding introduces two challenges. 
First, tokens from different beams may inadvertently attend to one another, resulting in corrupted outputs. 
Second, eliminated tokens may remain in memory, contradicting the goal of efficient memory usage. 
To address these issues, we adapt the attention mechanism to isolate branch-specific contexts and employ dynamic pruning to remove low-probability branches, ensuring both correctness and memory efficiency. 
These innovations enable our approach to achieve substantial memory savings while maintaining inference speed, offering a scalable solution for deploying LLMs in resource-constrained settings.

We conduct experiments to evaluate our trie-based decoding approach against greedy decoding and conventional batch-based beam search across three attention variants and two  datasets, showing the correctness and robustness of our method. 
The results demonstrate that our method achieves comparable performance to batch-based beam search with the same beam width, while substantially reducing memory usage, particularly for larger beam widths. 
Our contributions are as follows:
\begin{enumerate}
\item We propose a trie-based decoding method that significantly reduces memory usage during beam search by consolidating KV caches among beams with common prefixes, effectively addressing a critical limitation of batch-based beam search.

\item Under dense attention, our approach is theoretically equivalent to conventional beam search while substantially reducing memory overhead. 
Empirical results across three transformer architectures, Multi-Head, Grouped Query, and Sliding Window Attention, demonstrate that it preserves output quality, with differences from conventional beam search being statistically insignificant.

\item We release our implementation,\footnote{\url{https://github.com/brian030128/tridecode}} offering a scalable and practical decoding alternative. 
Unlike greedy decoding or top-$k$ sampling, our method retains beam search's robustness with significantly lower computational overhead, enabling efficient deployment of LLMs.
\end{enumerate}

\section{Related Work}
The evolution of decoding methods for language models in natural language processing (NLP) has been a subject of extensive research, focusing on improving both efficiency and output quality. This section reviews key developments in decoding strategies, including beam search, sampling methods, hybrid approaches, and recent advancements in computational efficiency.

The work of \citet{bahdanau2016neuralmachinetranslationjointly} marked a pivotal moment in NLP, introducing the attention mechanism, which allowed models to dynamically focus on relevant parts of the input sequence during generation. This breakthrough significantly improved translation quality, especially for long sentences. Notably, the study employed beam search as its decoding method—a technique that had already gained traction in statistical machine translation.

Following this milestone, beam search became the dominant decoding method, as evidenced by its use in prominent works like \citet{NIPS2017_3f5ee243} and \citet{wu2016googlesneuralmachinetranslation}. 
Beam search's ability to maintain multiple hypotheses during decoding often resulted in outputs that were more coherent and grammatically accurate, solidifying its popularity in constrained tasks.


Comparative analyses of decoding strategies have highlighted the trade-offs between beam search and sampling. \citet{ippolito-etal-2019-comparison} showed that while sampling methods generate more diverse outputs, they often compromise coherence and factual accuracy. Similarly, \citet{massarelli2020decodingstrategiesaffectverifiability} emphasized the susceptibility of sampling to hallucinations, contrasting this with beam search's strengths in accuracy and faithfulness, albeit at the cost of diversity. These trade-offs have inspired hybrid approaches, such as combining initial sampling with beam search~\citep{massarelli2020decodingstrategiesaffectverifiability}, to leverage the strengths of both methods.

Computational efficiency has become increasingly critical with the growth of model size and complexity. 
For example, \citet{NIPS2017_3f5ee243} proposed a high-level algorithm that reduces the beam search space by bounding the length deviation, improving both memory efficiency and speed without sacrificing output quality. 
While our method operates at a lower level, it can integrate with such techniques to further optimize memory usage.

The issue of prefix overlap has also been studied in speculative decoding. SpecInfer~\citep{Miao_2024}, \citet{spector2023accelerating}, and Medusa~\citep{10.5555/3692070.3692273} introduced tree-structured draft-token approaches with tree attention to improve efficiency. \citet{Qin_He_Prakriya_Cong_Sun_2025} further enhanced SpecInfer with dynamic beam width.

In contrast, speculative decoding focuses on accelerating local sampling, whereas our work reduces the memory footprint of global beam search. These are complementary directions, and our optimized beam search could even be integrated into speculative decoding frameworks to improve draft-token tree generation.


A trie, or prefix tree, is a tree-based data structure designed for efficient storage and retrieval of strings based on their prefixes~\citep{Briandais1959FileSU,10.1145/367390.367400}. 
A trie represents common prefixes of strings as shared nodes, enabling compact storage and efficient traversal.
Each node corresponds to a character, and the path from the root to any node represents a prefix of the stored string.

In the context of NLP, tries have been employed in tasks such as language modeling, dictionary construction, and decoding. 
Their ability to compactly represent shared prefixes makes them particularly suitable for beam search, where multiple beams often share a large number of overlapping prefixes. 
As decoding progresses, most beams converge on a dominant path, leading to substantial redundancy in the KV cache across different beams.

Our work leverages the trie structure to address this redundancy. 
By organizing beams into a trie, we consolidate overlapping prefixes into a single representation, significantly reducing memory usage. This trie-based approach ensures efficient storage of shared contexts while maintaining the integrity of independent beams during decoding. 
It highlights the natural fit of trie for optimizing beam search in LLMs, where memory constraints and computational efficiency are critical.

\section{Methodology}
This section introduces a trie-based decoding approach that addresses inefficiencies by consolidating overlapping prefixes into a shared representation, significantly reducing memory usage while maintaining comparable performance to traditional beam search. 
We outline the conventional batch-based beam search process, explain the proposed trie-based approach, and detail key innovations such as tree-based attention masking and garbage collection for efficient memory management.
\subsection{Batch-Based Beam Search}
The high-level concept of batch-based beam search is outlined in Algorithm~\ref{alg:std-beam-search}. 
In transformer-based token generation, each newly generated token attends to the KV cache of previously generated tokens. 
Due to the nature of matrix operations in attention mechanisms, all tokens within a sequence must share consistent hidden state dimensions. 
Consequently, in batched beam search, each beam maintains a separate and complete context KV cache to preserve distinct dimensional spaces.

During beam search, most candidate beams are eliminated early on as their cumulative probability scores fall outside the beam width $b$.
As decoding progresses, new branches predominantly grow from the single dominant path. This leads to significant redundancy, with multiple beams sharing overlapping prefixes, as illustrated in Figure~\ref{fig:bs-comparison}.

\begin{algorithm}[!t]
\caption{Standard Batch-Based Beam Search} \label{alg:std-beam-search}
\begin{algorithmic}[1]
\REQUIRE LLM in batch inference \( P(\mathbf{x}_{\text{batch}}|\mathbf{X}_{\text{batch}}) \), beam width \( b \), prompt \( x_1, \dots, x_t \), and target sequence length \( T \)
\STATE Initialize beam \( B_0 \gets \{(x_1, \dots, x_t)\} \)
\STATE Initialize empty KV cache \( \text{kv} \)
\FOR{\( i = t, \dots, T-1 \)}
    \STATE Stack all sequences in \( B_i \) into a batch tensor \( \mathbf{X}_{\text{batch}} \)
    \STATE Compute probabilities and update KV cache: \( \hat{B}_{i+1}, \text{kv} \gets P(\mathbf{x}_{\text{batch}} | \mathbf{X}_{\text{batch}}, \text{kv}) \)
    \STATE Select top \( b \) sequences: \( B_{i+1} \gets \texttt{top-b}(\hat{B}_{i+1}) \)
\ENDFOR
\STATE \RETURN Sequence in \( B_T \) with the highest cumulative probability
\end{algorithmic}
\end{algorithm}

\subsection{Trie-Based Decoding}
Our trie-based decoding approach leverages this redundancy by merging all branches with shared prefixes into a single dimension using a prefix tree traversal. 
This eliminates the need to store duplicated tokens across beams, significantly reducing memory usage.

However, this approach introduces two challenges. 
First, if the merged tensor is directly processed by the language model, tokens from different branches could attend to each other, corrupting the outputs. 
Second, eliminated tokens would persist in the tensor, unnecessarily occupying memory and undermining the goal of memory conservation. 
The following subsections present our solutions to these challenges.


\begin{algorithm}[t!]
\caption{Trie-Based Beam Search} \label{alg:trie-beam-search}
\begin{algorithmic}[1]
\REQUIRE LLM with trie attention \( P(x | \mathcal{T}, M) \), where \( \mathcal{T} \) is the trie structure, \( M \) is the attention mask corresponding to \( \mathcal{T} \), and \( x \) is the next token to predict; beam width \( b \), target sequence length \( L \), prompt, garbage collection interval \( g \)
\STATE Initialize a trie \( \mathcal{T} \gets \text{initialize\_trie}(\text{prompt}) \)
\STATE Initialize attention mask \( M \)
\STATE Serialize the trie to input: \( \text{input} \gets \texttt{serialize}(\mathcal{T}) \)
\FOR{\( i = |\text{input}|, \ldots, L-1 \)}
    \IF{\( i \bmod g = 0 \)}
        \STATE \texttt{garbage\_collect()} 
        \STATE \( M \gets \texttt{recompute\_mask}(\mathcal{T}, b) \)
    \ENDIF
    \STATE Predict the $b$ best tokens to expand $\mathcal{T}$: \( \text{V} \gets \mathrm{argsort}_{b} P(x | \text{input}, M) \)
    \STATE \( \mathcal{T} \gets \texttt{update\_trie}(\mathcal{T}, V) \)
    \STATE \( M \gets \texttt{update\_mask}(\mathcal{T}, M) \)
    \STATE Serialize updated trie for next iteration: \( \text{input} \gets \texttt{serialize}(\mathcal{T}) \)
\ENDFOR
\STATE \RETURN Sequence in \( \mathcal{T} \) with the highest cumulative probability
\end{algorithmic}
\end{algorithm}

\subsection{Tree Attention for Trie-based Decoding} \label{sec:tree-attention}

\begin{figure}[tbh]
    \centering
    \includegraphics[width=1\linewidth]{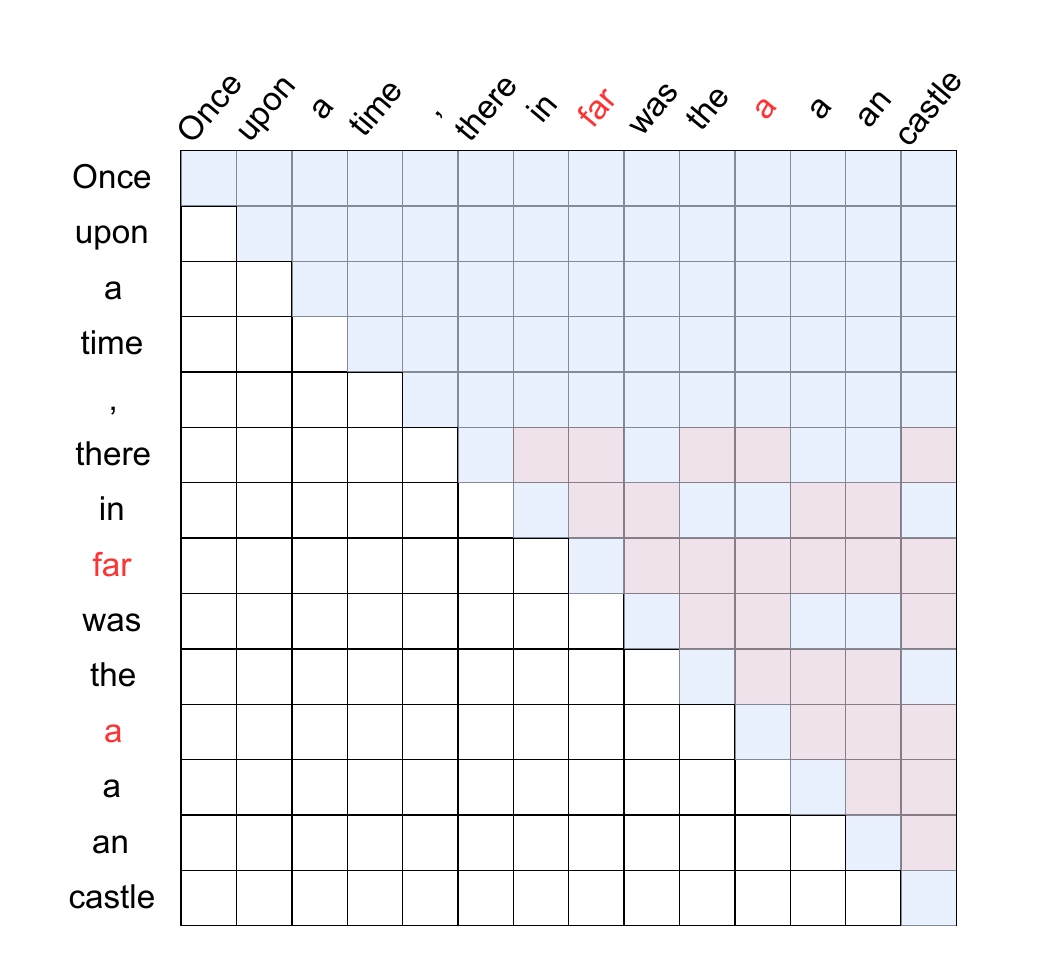}
    \caption{Causal attention masking that mirrors the trie structure. Rows and columns represent tokens in the trie. Blue cells indicate valid self-attention within a branch, while pink cells (masks) block cross-branch connections, preserving branch isolation.}
    \label{fig:mask}
\end{figure}

\begin{algorithm}[ht]
\caption{Causal Mask Construction for the Trie Structure} \label{alg:mask-construction}
\begin{algorithmic}[1]
\REQUIRE Trie \( \mathcal{T} \), input length \( t \), beam width \( b \)
\STATE Initialize attention mask \( M \in \mathbb{R}^{b \times (t + |\mathcal{T}|)} \gets -\infty \) \COMMENT{Initialize mask with negative infinity}
\STATE \( M[:, :t] \gets 0 \) \COMMENT{Allow attention to the input sequence}
\STATE Initialize temporary nodes: \( \mathbf{V} \gets \texttt{leaf\_nodes}(\mathcal{T}) \)
\WHILE{true}
    \STATE \( \text{reached\_root} \gets \text{true} \)
    \FOR{\( i = 1 \) to \( b \)}
        \STATE \( M[i, \texttt{idx}(V_i) + t] \gets 0 \) \COMMENT{Allow attention to current node}
        \IF{\( \texttt{parent}(V_i) \neq \emptyset \)}
            \STATE \( V_i \gets \texttt{parent}(V_i) \) \COMMENT{Move up the tree}
            \STATE \( \text{reached\_root} \gets \text{false} \) \COMMENT{Continue traversal}
        \ENDIF
    \ENDFOR
    \IF{\( \text{reached\_root} \)}
        \RETURN \( M \)
    \ENDIF
\ENDWHILE
\end{algorithmic}
\end{algorithm}

Combining multiple sequence branches into a single dimension improves processing efficiency but risks unwanted cross-branch interactions. 
For example, in Figure~\ref{fig:bs-comparison}, tokens like ``castle'' and ``was,'' which belong to different branches, should not influence one another during attention operations. 
To ensure branch independence, we construct a specialized causal attention mask that mirrors the structure of a trie, as detailed in Algorithm~\ref{alg:mask-construction}. 

During the attention mechanism, masks are applied by assigning large negative values to specific attention scores prior to the softmax operation. This ensures that the masked positions receive zero attention weight, thereby eliminating their influence on isolated branches. 
This standard transformer practice ensures isolation via masking. 
Since attention weights are computed relatively during the softmax phase, applying a mask beforehand effectively isolates cross-branch tokens without introducing interference. 

As illustrated in Figure~\ref{fig:mask}, this mask enforces that tokens attend only to other tokens within their respective branches, maintaining the integrity of each beam during the decoding process.
The attention mask is dynamically updated in two key steps. 
First, after selecting the top $b$
tokens at each decoding step, we update the mask to reflect the relationships between these tokens and their parent branches. 
Second, following garbage collection (Section~\ref{sec:gc}), we update the mask to account for the changes in the KV cache, ensuring consistency with the updated tree structure. 

\subsection{Maintenance of Positional Integrity}
In the trie-based beam search illustration, the renumbered position IDs are designed to simulate the exact behavior of conventional beam search, where the position ID of each token plays a crucial role in contextual understanding within the transformer architecture. Unlike conventional beam search, where each beam maintains independent position IDs, our trie-based approach merges shared prefixes and consolidates the tokens from all beams into a single structure. 
However, neighboring tokens within the same branch may be separated by tokens from other branches in this shared structure.

To preserve the integrity of positional information, the renumbered position IDs in our approach are assigned to match the positions in the original beam search. 
This ensures that the contextual dependency between tokens in the same branch remains intact, even when tokens from different branches are interleaved. 
By aligning the position IDs with those in the conventional beam search, our method achieves equivalent contextual understanding while maintaining memory efficiency. 
This alignment is critical for ensuring that the model generates outputs consistent with the original beam search behavior, while leveraging the benefits of the trie structure to reduce redundancy.

\subsection{Garbage Collection} \label{sec:gc}
Garbage collection (GC) consolidates and reclaims unused memory. Because GPU memory operations are expensive, we minimize overhead by deferring token removal and KV cache reorganization. Instead of updating at every decoding step, we accumulate changes and trigger garbage collection only after a predefined threshold is reached. The procedure executes in three stages:

\begin{enumerate}
    \item \textbf{Marking}: Traverse the tree bottom-up from leaf nodes to the root, marking all unvisited nodes for removal (CPU).
    \item \textbf{Pruning}: Eliminate marked nodes from the CPU-side reference structure via a lightweight traversal (CPU).
    \item \textbf{Compaction}: Compact the KV cache by discarding marked tokens, using \texttt{torch.index\_select} to retain only unmarked entries (GPU).
\end{enumerate}

At each GC, we reconstruct the decoding sequence from the surviving nodes. 
This design follows the amortized philosophy of scapegoat trees~\citep{galperin1993scapegoat}: rather than paying incremental maintenance costs at every step, we periodically rebuild the structure in linear time, achieving predictable long-term efficiency.



\section{Experiments}
\begin{table*}[!tbh]
  \centering
  \setlength{\tabcolsep}{2pt}
  \renewcommand\theadfont{\normalsize\bfseries}
  \resizebox{\textwidth}{!}{%
  \begin{tabular}{llc
                  rrr
                  rrr
                  rr
                  rr}
    \toprule
    \multirow{2}{*}{\textbf{Model}} &
    \multirow{2}{*}{\textbf{Dataset}} &
    \multirowcell{2}{\textbf{Beam}} &
    \multicolumn{3}{c}{\textbf{Mem/Tok (MB)}\,$\downarrow$} &
    \multicolumn{3}{c}{\textbf{Tok/Sec}\,$\uparrow$} &
    \multicolumn{2}{c}{\textbf{Score (mean)}} &
    \multicolumn{2}{c}{\textbf{Gain (×)}} \\
    \cmidrule(lr){4-6}\cmidrule(lr){7-9}\cmidrule(lr){10-11}\cmidrule(lr){12-13}
     & & &
     Origin & Trie & $\Delta\!\pm$CI &
     Origin & Trie & $\Delta\!\pm$CI &
     Origin & Trie &
     Mem. & Speed \\
    \midrule
\multirow{7}{*}{\textbf{Llama 3.1 8B}} & \multirow{3}{*}{CNN} & 1  & 0.32 & \textit{N/A} & \textit{N/A} & 5.32 & \textit{N/A} & \textit{N/A} & 0.20 & \textit{N/A} & \textit{N/A} & \textit{N/A} \\
                &     & 3  & 0.85 & 0.34 & $-0.51^{\dagger}\!\pm\!0.00$ & 4.45 & 10.96 & $+6.51^{\dagger}\!\pm\!0.13$ & 0.21 & 0.20 & $2.50\times$ & $2.46\times$ \\
                &     & 9  & 2.45 & 0.43 & $-2.02^{\dagger}\!\pm\!0.01$ & 3.69 & 10.02 & $+6.33^{\dagger}\!\pm\!0.18$ & 0.20 & 0.20 & $5.70\times$ & $2.72\times$ \\
\cmidrule(lr){2-13}
                & \multirow{4}{*}{HumanEval} & 1  & 0.45 & \textit{N/A} & \textit{N/A} & 5.61 & \textit{N/A} & \textit{N/A} & 0.60 & \textit{N/A} & \textit{N/A} & \textit{N/A} \\
                &              & 3  & 1.01 & 0.58 & $-0.43^{\dagger}\!\pm\!0.01$ & 4.98 & 12.53 & $+7.55^{\dagger}\!\pm\!0.05$ & 0.65 & 0.65 & $1.74\times$ & $2.52\times$ \\
                &              & 9  & 2.67 & 0.85 & $-1.82^{\dagger}\!\pm\!0.02$ & 4.41 & 11.81 & $+7.40^{\dagger}\!\pm\!0.10$ & 0.66 & 0.65 & $3.14\times$ & $2.68\times$ \\
                &              & 15 & 4.35 & 1.09 & $-3.26^{\dagger}\!\pm\!0.04$ & 4.04 & 11.23 & $+7.19^{\dagger}\!\pm\!0.15$ & 0.65 & 0.65 & $4.00\times$ & $2.78\times$ \\
\midrule
\multirow{6}{*}{\textbf{Mistral-Small}} & \multirow{3}{*}{CNN} & 1  & 0.37 & \textit{N/A} & \textit{N/A} & 7.23 & \textit{N/A} & \textit{N/A} & 0.17 & \textit{N/A} & \textit{N/A} & \textit{N/A} \\
                &     & 3  & 1.04 & 0.41 & $-0.63^{\dagger}\!\pm\!0.01$ & 5.52 & 6.56 & $+1.04^{\dagger}\!\pm\!0.07$ & 0.17 & 0.17 & $2.54\times$ & $1.19\times$ \\
                &     & 6  & 2.04 & 0.49 & $-1.55^{\dagger}\!\pm\!0.01$ & 4.13 & 5.74 & $+1.61^{\dagger}\!\pm\!0.10$ & 0.16 & 0.16 & $4.16\times$ & $1.39\times$ \\
\cmidrule(lr){2-13}
                & \multirow{3}{*}{HumanEval} & 1  & 0.52 & \textit{N/A} & \textit{N/A} & 4.32 & \textit{N/A} & \textit{N/A} & 0.74 & \textit{N/A} & \textit{N/A} & \textit{N/A} \\
                &              & 3  & 1.21 & 0.68 & $-0.53^{\dagger}\!\pm\!0.02$ & 3.71 & 7.32 & $+3.61^{\dagger}\!\pm\!0.03$ & 0.78 & 0.78 & $1.78\times$ & $1.97\times$ \\
                &              & 6  & 2.23 & 0.86 & $-1.37^{\dagger}\!\pm\!0.02$ & 3.33 & 6.59 & $+3.26^{\dagger}\!\pm\!0.06$ & 0.77 & 0.77 & $2.59\times$ & $1.98\times$ \\
\midrule
\multirow{7}{*}{\textbf{Phi-3.5-mini}} & \multirow{3}{*}{CNN} & 1  & 1.36 & \textit{N/A} & \textit{N/A} & 14.13 & \textit{N/A} & \textit{N/A} & 0.19 & \textit{N/A} & \textit{N/A} & \textit{N/A} \\
                 &     & 3  & 4.00 & 1.39 & $-2.61^{\dagger}\!\pm\!0.11$ & 11.51 & 10.81 & $-0.70^{\dagger}\!\pm\!0.07$ & 0.20 & 0.19 & $2.88\times$ & $0.94\times$ \\
                 &     & 9  & 11.95 & 1.39 & $-10.56^{\dagger}\!\pm\!0.19$ & 4.23 & 8.88 & $+4.65^{\dagger}\!\pm\!0.09$ & 0.19 & 0.19 & $8.59\times$ & $2.10\times$ \\
\cmidrule(lr){2-13}
                 & \multirow{4}{*}{HumanEval} & 1  & 1.11 & \textit{N/A} & \textit{N/A} & 13.79 & \textit{N/A} & \textit{N/A} & 0.65 & \textit{N/A} & \textit{N/A} & \textit{N/A} \\
                 &              & 3  & 3.23 & 1.27 & $-1.96^{\dagger}\!\pm\!0.08$ & 12.82 & 13.99 & $+1.17^{\dagger}\!\pm\!0.10$ & 0.69 & 0.69 & $2.55\times$ & $1.09\times$ \\
                 &              & 9  & 9.63 & 2.21 & $-7.42^{\dagger}\!\pm\!0.29$ & 10.89 & 11.86 & $+0.97^{\dagger}\!\pm\!0.10$ & 0.69 & 0.70 & $4.36\times$ & $1.09\times$ \\
                 &              & 15 & 15.97 & 3.11 & $-12.86^{\dagger}\!\pm\!0.46$ & 9.85 & 10.96 & $+1.11^{\dagger}\!\pm\!0.13$ & 0.70 & 0.70 & $5.14\times$ & $1.11\times$ \\
    \bottomrule
  \end{tabular}}
\caption{Comparison of our trie-based decoding vs.\ conventional beam search. 
Means over 1,000 samples; $\Delta = \text{Trie} - \text{Origin}$. 
Statistical significance ($p<0.01$) for efficiency deltas is indicated by $\dagger$. 
\textbf{Score} reports no significant difference in quality metrics (ROUGE-L for CNN; Accuracy for HumanEval) across all models, datasets, and beam widths. 
\textbf{Memory Efficiency Gain} = Origin\,/\,Trie; \textbf{Speed Efficiency Gain} = Trie\,/\,Origin.
For beam size = 1 (greedy decoding), both methods are identical; thus, comparison cells are marked \textit{N/A}.
}\label{tab:trie_beam_results}
\end{table*}

\subsection{Experimental Setup}
We evaluated our trie-based decoding on three representative transformer models chosen to demonstrate the generalizability of our approach across popular attention mechanisms:
Multi-Head Attention (Phi-3.5-mini-instruct; \citealp{abdin2024phi3technicalreporthighly}),
Grouped Query Attention (Llama-3.1-8B-Instruct\footnote{\url{https://github.com/meta-llama/llama-models/blob/main/models/llama3_1/MODEL_CARD.md}}),
and Sliding Window Attention (Mistral-Small-24B-Instruct-2501\footnote{\url{https://huggingface.co/mistralai/Mistral-Small-24B-Instruct-2501}}).

Experiments were conducted on two diverse generation tasks:
abstractive summarization (CNN/DailyMail dataset; \citealp{nallapati-etal-2016-abstractive}) evaluated using ROUGE-L scores~\citep{lin2004rouge}, 
and code generation (HumanEval dataset; \citealp{chen2021evaluatinglargelanguagemodels}) evaluated using binary accuracy.

In addition to generation quality, we evaluated two efficiency metrics.
Memory efficiency was defined as memory consumption per processed token:

\[
\frac{\text{Peak memory} - \text{Model memory}}{\text{Input length} + \text{Output length}}
\]

\noindent Here, ``Model memory'' denotes the fixed GPU memory required to load the model, which is excluded from comparisons. 
Decoding speed was measured in tokens per second, computed as the ratio of output length to inference time. 

All evaluations were conducted in a one-shot setting across beam widths ($b$). 
Experiments employed four Tesla V100-SXM2-32GB GPUs for Llama 3.1 and Mistral, and a single GPU for Phi-3.5, demonstrating both single- and multi-GPU compatibility of our algorithm. 
Statistical significance was assessed using paired \textit{t}-tests for ROUGE-L, memory efficiency, and decoding speed, and the McNemar test for Accuracy.

\subsection{Output Fidelity}\label{sec:quality}
Table~\ref{tab:trie_beam_results} summarizes the experimental results.
Despite substantial efficiency gains, our trie-based decoding is intended to be mathematically equivalent to conventional beam search. 
Across beam widths, it achieves nearly identical ROUGE-L scores on CNN and comparable accuracies on HumanEval, with no statistically significant differences ($p < 0.01$). 
Minor variations arise from implementation details, numerical precision, and the lack of batch invariance~\citep{he2025nondeterminism}.

A further limitation stems from sparse attention mechanisms~\citep{child2019generatinglongsequencessparse}: because they modify the effective dependency structure, the flattened trie may not perfectly replicate the behavior of dense-attention beam search. This structural divergence can introduce additional discrepancies, albeit typically small in practice.

To further validate fidelity, we compared per-token logit distributions from trie-based decoding \(T_t \in \mathbb{R}^{B \times V}\) and batch-based beam search \(B_t \in \mathbb{R}^{B \times V}\) at each decoding step $t$. 
Logits were evaluated up to the first divergence point in the decoding tree. 
At $b=3$ across all models and datasets, the average softmax-normalized differences remained below $10^{-5}$, effectively at machine precision, confirming equivalence in behavior. 

These complementary analyses confirm that trie-based decoding reproduces the outputs of conventional beam search to machine precision in practice. Both the sequence-level results in Table~\ref{tab:trie_beam_results} and the per-token logit comparisons show outputs that are indistinguishable across all tested settings. Sparse attention mechanisms may theoretically alter the dependency structure and limit strict equivalence, but such effects did not manifest empirically in our experiments. Overall, trie-based decoding provides a faithful and efficient alternative to beam search, while sparse-attention–aware extensions remain a direction for future work.

\subsection{Efficiency Analysis}

\begin{figure*}[!tbh]
    \centering
    \includegraphics[width=1\linewidth]{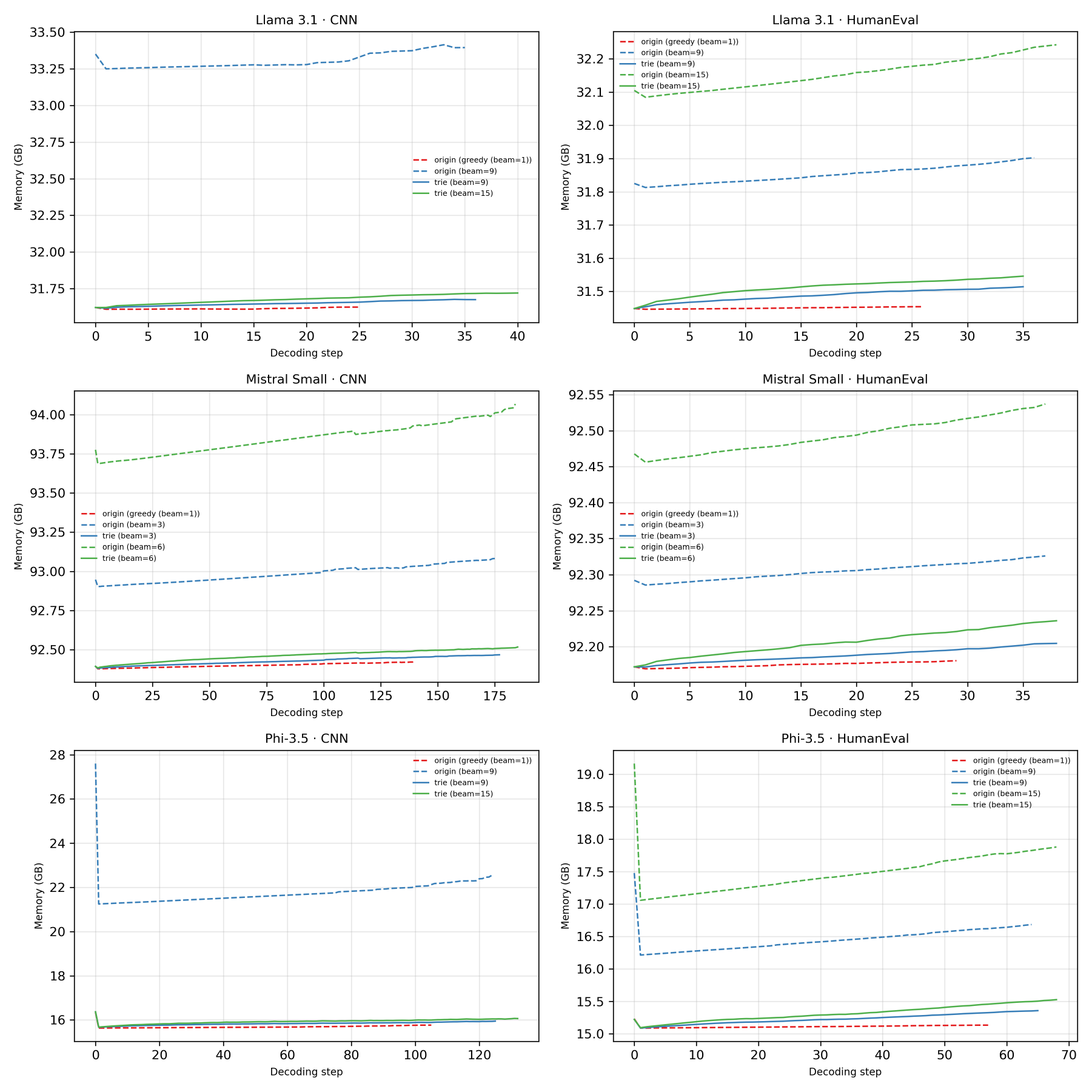}
    \caption{Memory usage during decoding across three models and tasks. Dashed lines show standard beam search; solid lines show trie-based decoding. Each point is the average at a decoding step, truncated when fewer than 80\% of samples remain. Our trie-based decoding consistently reduces memory usage, closely matching greedy decoding.}
    \label{fig:memory_comparison}
\end{figure*}

\paragraph{Memory Efficiency}
As shown in Table~\ref{tab:trie_beam_results}, our trie-based decoding substantially reduces memory usage across all models and beam widths. 
For larger beam widths (e.g., 9 or 15), we observe memory savings of 4--8 times for Phi-3.5-mini and 4--6 times for Llama 3.1 and Mistral-Small. 
Notably, our method achieves memory usage comparable to greedy decoding (beam width of 1), as illustrated in Figure~\ref{fig:memory_comparison}, highlighting its suitability for deployment in memory-constrained environments.

\paragraph{Time Efficiency}
Table~\ref{tab:trie_beam_results} also confirms that our approach consistently improves decoding speed, especially at larger beam widths. For instance, Phi-3.5-mini achieves a speedup of $2.42\times$ at beam width 9, while Mistral attains $1.38\times$ at beam width 6. 
Although speed was not our primary optimization goal, these improvements are significant, highlighting reduced memory transfer overhead and further enhancing practical applicability.
 
As summarized in Table~\ref{tab:trie_beam_results}, multiplicative gains in memory (Memory Gain = Origin\,/\,Trie) and speed (Speed Gain = Trie\,/\,Origin) consistently increase with wider beams, 
highlighting the scalability and robustness of trie-based decoding. 
Although optimal beam width remains task-dependent, our results show that it preserves the output quality of conventional beam search while substantially improving efficiency.

\section{Discussion}

\begin{table}[t!]
\centering
\setlength{\tabcolsep}{3pt}
\begin{tabular}{lccc}
\toprule
\textbf{Configuration} & \textbf{Trie} & \textbf{GC} & \textbf{Saved Tokens} \\
\midrule
Original Beam Search         &  &  & $0.0 \pm 0.0$ \\
Trie-based  w/o GC                  & \checkmark &  & $360.3 \pm 22.1$ \\
Trie-based (Ours)             & \checkmark & \checkmark & $535.8 \pm 32.3$ \\
\bottomrule
\end{tabular}
\caption{Results of the ablation study on Phi-3.5-mini with beam width of
3 on HumanEval. ``Saved Tokens'' denotes the average number of KV cache entries avoided.}
\label{tab:ablation}
\end{table}

\subsection{Ablation Study}
We performed an ablation study to evaluate the individual contributions of the two core components in our approach:  
(1) the \textbf{trie-based attention masking}, which removes redundant tokens by consolidating shared prefixes; and  
(2) the \textbf{garbage collection mechanism}, which reclaims memory by eliminating obsolete branches from the KV cache.

Experiments were conducted using the Phi-3.5-mini model on the HumanEval dataset with a beam width of 3. The results are summarized in Table~\ref{tab:ablation}.
The trie-based attention masking alone 
yields substantial savings of approximately 360 tokens per run in KV cache storage, confirming its effectiveness in mitigating redundancy from shared prefixes.
However, without GC, obsolete branches persist in memory, limiting scalability. Incorporating GC further improves efficiency, reaching an average saving of 536 tokens, and consistently prevents accumulation of unused branches. These results demonstrate that the trie reduces redundancy, while GC sustains efficiency. 

\subsection{Memory Usage During Decoding}
As shown in Figure~\ref{fig:memory_comparison}, all methods exhibit a temporary memory spike during the prefilling stage, most noticeable for the Phi-3.5-mini model. 
The spike originates from a large intermediate output tensor of size (beam width $\times$ input length $\times$ vocab size) generated during prefilling.   
Although this tensor is released immediately afterward, the spike is disproportionately pronounced in Phi-3.5-mini due to its smaller model size, accentuating its relative impact on overall memory usage.

\subsection{Overhead of Garbage Collection}

\begin{figure}[!t]
    \centering
    \includegraphics[width=1\linewidth]{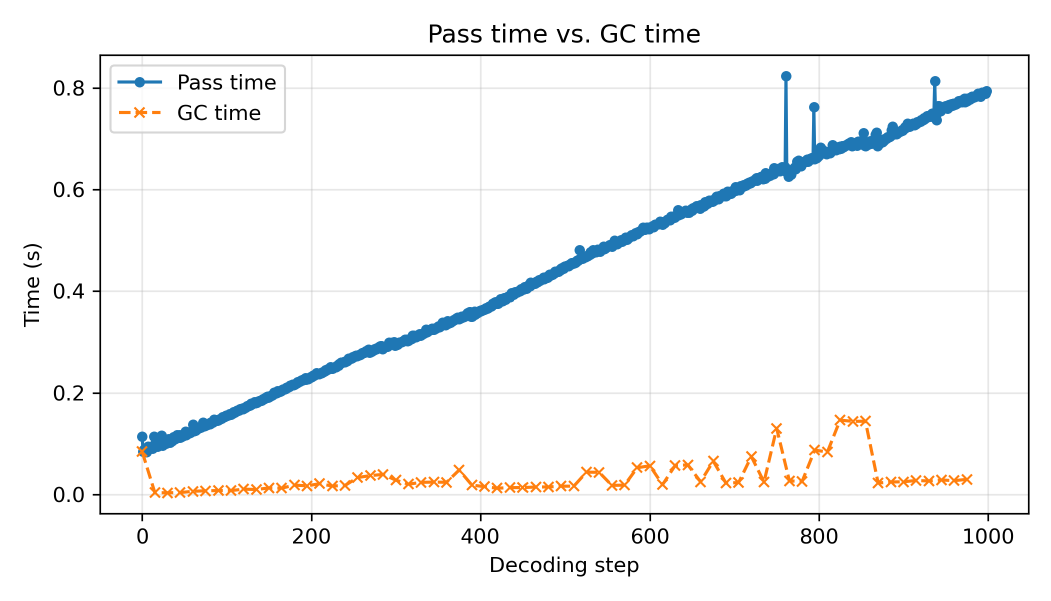}
    \caption{Comparison between GC time and decoding pass time across 1,000 decoding steps with a beam width of 30. 
    GC is triggered every 15 decoding steps in this experiment. 
    The results show that GC overhead remains low and scales more favorably than the decoding pass time, which increases steadily over time.}
    \label{fig:gc_time_comparison}
\end{figure}

We conducted an experiment on Llama 3.1 8B using 30 beams to generate 1,000 tokens, as shown in Figure~\ref{fig:gc_time_comparison}. 
GC consumed less than 20\% of the time required for a single decoding pass and was further amortized over 15 decoding steps, resulting in negligible impact on overall decoding time. 
While the GC overhead does increase with the number of generated tokens, it scales more slowly than the decoding time. 
This is primarily because attention operations scale quadratically,  whereas GC scales linearly. 
Although CPU and GPU operation costs are not directly comparable, this observation provides insight into the relative insignificance of GC overhead. 
Given this negligible overhead, we did not perform further optimization, though additional refinements remain possible.

\subsection{Trie-based Decoding for Reasoning}
We evaluate our trie-based beam search on reasoning-heavy tasks using the MATH500 dataset with chain-of-thought (CoT) prompting (``think step by step'')~\citep{hendrycks2021measuring,lightman2023let}. 
As baselines, we compare against top-$k$ sampling ($k=3$), a standard approach for reasoning tasks, and greedy decoding, a special case of top-$k$ sampling ($k=1$) and of beam search (beam width $b=1$).

Table~\ref{tab:math500_results} shows that trie-based decoding consistently improves accuracy as beam width increases, outperforming both greedy decoding and top-$k$ sampling.
For example, accuracy rises from 0.23 (Llama 3.1 8B) and 0.41 (Phi-3.5-mini) under top-$k$ sampling to 0.40 and 0.47 with a beam width of 15. 
These results confirm that wider beams enhance reasoning consistency and correctness.
We also observe that beam search tends to produce longer CoT sequences, consistent with prior findings on the positive correlation between reasoning length and accuracy~\citep{guo2025deepseek}.

In terms of efficiency, greedy decoding and top-$k$ sampling use a single beam, resulting in lower memory usage and faster decoding.
By design, trie-based decoding maintains multiple beams, increasing memory consumption and reducing throughput. 
This gap is expected and not a direct comparison. Nevertheless, the overhead remains modest and practically tolerable.

\begin{table}[!t]
  \centering
  \setlength{\tabcolsep}{3pt}
  \begin{tabular}{lcrrc}
\multicolumn{5}{c}{\textbf{Results on Llama 3.1 8B}} \\
    \toprule
    \textbf{Method} &
    \textbf{Mem/Tok} &
    \textbf{Tok/Sec} &
    \textbf{Acc.} &
    \textbf{$\overline{\ell}_{out}$} \\
    \midrule
    Greedy & 0.28 & 14.03 & 0.30 & 411.98 \\
    Top-$3$   & 0.29 & 14.08 & 0.23 & 383.68 \\
    Trie ($b=3$)           & 0.38 & 12.13 & 0.33 & 421.80\\
    Trie ($b=9$)           & 0.68 &  9.59 & 0.38 & 453.68 \\
    Trie ($b=15$)          & 1.00 &  8.22 & 0.40 & 440.44 \\
    \bottomrule
    \\
    \multicolumn{5}{c}{\textbf{Results on Mistral-Small 24B}} \\
    \toprule
    \textbf{Method} &
    \textbf{Mem/Tok} &
    \textbf{Tok/Sec} &
    \textbf{Acc.} & 
    \textbf{$\overline{\ell}_{out}$} \\
    \midrule
    Greedy & 0.18 & 8.38 & 0.62 & 434.63 \\
    Top-$3$   & 0.18 & 8.21 & 0.35 & 383.24 \\
    Trie ($b=3$)           & 0.30 & 7.90 & 0.63 & 434.78 \\
    Trie ($b=6$)           & 0.44 &  7.31 & 0.65 & 448.66 \\
    \bottomrule
    \\
    \multicolumn{5}{c}{\textbf{Results on Phi-3.5-mini 3.8B}} \\
    \toprule
    \textbf{Method} &
    \textbf{Mem/Tok} &
    \textbf{Tok/Sec} &
    \textbf{Acc.} & 
    \textbf{$\overline{\ell}_{out}$} \\
    \midrule
    Greedy & 0.81 & 12.36 & 0.43 & 277.25 \\
    Top-$3$ & 0.81 & 12.36 & 0.41 & 271.60 \\
    Trie ($b=3$)           & 1.22 & 10.93 & 0.44 & 290.95 \\
    Trie ($b=9$) & 2.37 &  8.74 & 0.43 & 306.74 \\
    Trie ($b=15$)          & 3.56 &  7.74 & 0.47 & 302.99 \\
    \bottomrule
  \end{tabular}
\caption{Comparison of greedy decoding, top-$k$ sampling, and trie-based beam search on MATH500 with CoT prompting.
Memory usage is measured in MB, and $\overline{\ell}_{\text{out}}$ denotes the average output length in tokens.}
\label{tab:math500_results}
\end{table}

\section{Conclusion}
This work presents trie-based decoding, a novel beam search method for LLMs that significantly reduces memory usage by consolidating shared prefixes among beams. 
Our approach demonstrates substantial improvements in memory efficiency, and its effectiveness has been validated across three popular modern transformer architectures, including Multi-Head, Grouped Query, and Sliding Window Attention. 

Trie-based decoding is especially beneficial for tasks with large contexts and wide beams, such as code generation. 
Our approach requires no additional training or specialized hardware, offering a practical, scalable, and cost-effective solution for deploying LLMs in resource-constrained settings.

\section*{Acknowledgments}
We thank the reviewers and area chair for their valuable suggestions that strengthened the final version. 
This research was partially supported by the National Science and Technology Council (NSTC), Taiwan, under Grant No. 112-2221-E-001-016-MY3; by Academia Sinica under Grant No. 236d-1120205; and by the National Center for High-performance Computing, National Applied Research Laboratories, and NSTC under the ``Trustworthy AI Dialog Engine (TAIDE)'' project.

\section*{Limitations}
While our trie-based decoding method significantly improves memory and decoding speed, it also has several limitations:

\begin{itemize}

\item \textbf{Interaction with Sparse Attention}: 
Sparse attention may modify the dependency structure, preventing strict equivalence. 
However, this effect did not manifest empirically in our evaluations. We therefore conclude that trie-based decoding offers a faithful and efficient substitute for conventional beam search in practice, while sparse-attention–aware extensions remain a promising avenue for future work.

\item \textbf{Garbage Collection Overhead}: 
Acceleration primarily stems from reduced memory usage. 
In modern LLMs, the computational bottleneck often lies in memory bandwidth, so when memory usage is high, our implementation tends to be faster than the original. 
However, garbage collection introduces some overhead. 
As a result, when the model size is small, the beam width is narrow, and the total number of tokens is low, our method may actually be slightly slower. 
This can be observed in the case of the small 3.8B Phi-3.5-mini model in Table~\ref{tab:trie_beam_results}.

\item \textbf{Task Dependency}:
Our approach offers substantial gains primarily when beams share common prefixes, making it most beneficial for tasks where beams frequently converge. 
Its efficiency gain may diminish for tasks involving highly diverse or divergent outputs.

\item \textbf{Evaluation Scope}: Our experiments focus on summarization (CNN/DailyMail) and code generation (HumanEval), evaluating three mainstream transformer architectures. While these results support the generalizability of our approach, its performance on other tasks or models, such as multimodal transformers or retrieval-augmented generation, requires further investigation.
\end{itemize}
Future work should address these limitations, exploring broader task applicability.

\section*{Ethical Considerations}
This work exclusively utilizes publicly available datasets (CNN/DailyMail and HumanEval), which contain no personally identifiable or sensitive information. 
Additionally, we disclose that the manuscript underwent minor language refinement and polishing using ChatGPT. 
The authors retain full responsibility for all content presented.

\bibliography{main}

\begin{thebibliography}{27}
\providecommand{\natexlab}[1]{#1}

\bibitem[{Abdin et~al.(2024)Abdin, Aneja, Awadalla, Awadallah, Awan, Bach, Bahree, Bakhtiari, Bao, Behl, Benhaim, Bilenko, Bjorck, Bubeck, Cai, Cai, Chaudhary, Chen, Chen, Chen, Chen, Chen, Cheng, Chopra, Dai, Dixon, Eldan, Fragoso, Gao, Gao, Gao, Garg, Giorno, Goswami, Gunasekar, Haider, Hao, Hewett, Hu, Huynh, Iter, Jacobs, Javaheripi, Jin, Karampatziakis, Kauffmann, Khademi, Kim, Kim, Kurilenko, Lee, Lee, Li, Li, Liang, Liden, Lin, Lin, Liu, Liu, Liu, Liu, Liu, Luo, Madan, Mahmoudzadeh, Majercak, Mazzola, Mendes, Mitra, Modi, Nguyen, Norick, Patra, Perez-Becker, Portet, Pryzant, Qin, Radmilac, Ren, de~Rosa, Rosset, Roy, Ruwase, Saarikivi, Saied, Salim, Santacroce, Shah, Shang, Sharma, Shen, Shukla, Song, Tanaka, Tupini, Vaddamanu, Wang, Wang, Wang, Wang, Wang, Wang, Ward, Wen, Witte, Wu, Wu, Wyatt, Xiao, Xu, Xu, Xu, Xue, Yadav, Yang, Yang, Yang, Yang, Yu, Yuan, Zhang, Zhang, Zhang, Zhang, Zhang, Zhang, Zhang, and Zhou}]{abdin2024phi3technicalreporthighly}
Marah Abdin, Jyoti Aneja, Hany Awadalla, Ahmed Awadallah, Ammar~Ahmad Awan, Nguyen Bach, Amit Bahree, Arash Bakhtiari, Jianmin Bao, Harkirat Behl, Alon Benhaim, Misha Bilenko, Johan Bjorck, Sébastien Bubeck, Martin Cai, Qin Cai, Vishrav Chaudhary, Dong Chen, Dongdong Chen, and 110 others. 2024.
\newblock \href {https://arxiv.org/abs/2404.14219} {Phi-3 technical report: A highly capable language model locally on your phone}.
\newblock \emph{Preprint}, arXiv:2404.14219.

\bibitem[{Bahdanau et~al.(2016)Bahdanau, Cho, and Bengio}]{bahdanau2016neuralmachinetranslationjointly}
Dzmitry Bahdanau, Kyunghyun Cho, and Yoshua Bengio. 2016.
\newblock \href {https://arxiv.org/abs/1409.0473} {Neural machine translation by jointly learning to align and translate}.
\newblock \emph{Preprint}, arXiv:1409.0473.

\bibitem[{Briandais(1959)}]{Briandais1959FileSU}
Rene De~La Briandais. 1959.
\newblock \href {https://api.semanticscholar.org/CorpusID:10963780} {File searching using variable length keys}.
\newblock In \emph{IRE-AIEE-ACM Computer Conference}.

\bibitem[{Cai et~al.(2024)Cai, Li, Geng, Peng, Lee, Chen, and Dao}]{10.5555/3692070.3692273}
Tianle Cai, Yuhong Li, Zhengyang Geng, Hongwu Peng, Jason~D. Lee, Deming Chen, and Tri Dao. 2024.
\newblock {MEDUSA}: Simple llm inference acceleration framework with multiple decoding heads.
\newblock In \emph{Proceedings of the 41st International Conference on Machine Learning}, ICML'24. JMLR.org.

\bibitem[{Chen et~al.(2021)Chen, Tworek, Jun, Yuan, de~Oliveira~Pinto, Kaplan, Edwards, Burda, Joseph, Brockman, Ray, Puri, Krueger, Petrov, Khlaaf, Sastry, Mishkin, Chan, Gray, Ryder, Pavlov, Power, Kaiser, Bavarian, Winter, Tillet, Such, Cummings, Plappert, Chantzis, Barnes, Herbert-Voss, Guss, Nichol, Paino, Tezak, Tang, Babuschkin, Balaji, Jain, Saunders, Hesse, Carr, Leike, Achiam, Misra, Morikawa, Radford, Knight, Brundage, Murati, Mayer, Welinder, McGrew, Amodei, McCandlish, Sutskever, and Zaremba}]{chen2021evaluatinglargelanguagemodels}
Mark Chen, Jerry Tworek, Heewoo Jun, Qiming Yuan, Henrique~Ponde de~Oliveira~Pinto, Jared Kaplan, Harri Edwards, Yuri Burda, Nicholas Joseph, Greg Brockman, Alex Ray, Raul Puri, Gretchen Krueger, Michael Petrov, Heidy Khlaaf, Girish Sastry, Pamela Mishkin, Brooke Chan, Scott Gray, and 39 others. 2021.
\newblock \href {https://arxiv.org/abs/2107.03374} {Evaluating large language models trained on code}.
\newblock \emph{Preprint}, arXiv:2107.03374.

\bibitem[{Child et~al.(2019)Child, Gray, Radford, and Sutskever}]{child2019generatinglongsequencessparse}
Rewon Child, Scott Gray, Alec Radford, and Ilya Sutskever. 2019.
\newblock \href {https://arxiv.org/abs/1904.10509} {Generating long sequences with sparse transformers}.
\newblock \emph{Preprint}, arXiv:1904.10509.

\bibitem[{Dao(2024)}]{dao2023flashattention2}
Tri Dao. 2024.
\newblock Flash{A}ttention-2: Faster attention with better parallelism and work partitioning.
\newblock In \emph{International Conference on Learning Representations (ICLR)}.

\bibitem[{Dao et~al.(2022)Dao, Fu, Ermon, Rudra, and R{\'e}}]{dao2022flashattention}
Tri Dao, Daniel~Y. Fu, Stefano Ermon, Atri Rudra, and Christopher R{\'e}. 2022.
\newblock Flash{A}ttention: Fast and memory-efficient exact attention with {IO}-awareness.
\newblock In \emph{Advances in Neural Information Processing Systems (NeurIPS)}.

\bibitem[{Dziri et~al.(2021)Dziri, Madotto, Za{\"i}ane, and Bose}]{dziri-etal-2021-neural}
Nouha Dziri, Andrea Madotto, Osmar Za{\"i}ane, and Avishek~Joey Bose. 2021.
\newblock \href {https://doi.org/10.18653/v1/2021.emnlp-main.168} {Neural path hunter: Reducing hallucination in dialogue systems via path grounding}.
\newblock In \emph{Proceedings of the 2021 Conference on Empirical Methods in Natural Language Processing}, pages 2197--2214, Online and Punta Cana, Dominican Republic. Association for Computational Linguistics.

\bibitem[{Fredkin(1960)}]{10.1145/367390.367400}
Edward Fredkin. 1960.
\newblock \href {https://doi.org/10.1145/367390.367400} {Trie memory}.
\newblock \emph{Commun. ACM}, 3(9):490–499.

\bibitem[{Galperin and Rivest(1993)}]{galperin1993scapegoat}
Igal Galperin and Ronald~L Rivest. 1993.
\newblock Scapegoat trees.
\newblock In \emph{Proceedings of the fourth annual ACM-SIAM Symposium on Discrete algorithms}, pages 165--174.

\bibitem[{Guo et~al.(2025)Guo, Yang, Zhang, Song, Wang, Zhu, Xu, Zhang, Ma, Bi et~al.}]{guo2025deepseek}
Daya Guo, Dejian Yang, Haowei Zhang, Junxiao Song, Peiyi Wang, Qihao Zhu, Runxin Xu, Ruoyu Zhang, Shirong Ma, Xiao Bi, and 1 others. 2025.
\newblock Deepseek-r1 incentivizes reasoning in llms through reinforcement learning.
\newblock \emph{Nature}, 645(8081):633--638.

\bibitem[{He and Lab(2025)}]{he2025nondeterminism}
Horace He and Thinking~Machines Lab. 2025.
\newblock \href {https://doi.org/10.64434/tml.20250910} {Defeating nondeterminism in llm inference}.
\newblock \emph{Thinking Machines Lab: Connectionism}.
\newblock Https://thinkingmachines.ai/blog/defeating-nondeterminism-in-llm-inference/.

\bibitem[{Hendrycks et~al.(2021)Hendrycks, Burns, Kadavath, Arora, Basart, Tang, Song, and Steinhardt}]{hendrycks2021measuring}
Dan Hendrycks, Collin Burns, Saurav Kadavath, Akul Arora, Steven Basart, Eric Tang, Dawn Song, and Jacob Steinhardt. 2021.
\newblock \href {https://openreview.net/forum?id=7Bywt2mQsCe} {Measuring mathematical problem solving with the {MATH} dataset}.
\newblock In \emph{Thirty-fifth Conference on Neural Information Processing Systems Datasets and Benchmarks Track (Round 2)}.

\bibitem[{Ippolito et~al.(2019)Ippolito, Kriz, Sedoc, Kustikova, and Callison-Burch}]{ippolito-etal-2019-comparison}
Daphne Ippolito, Reno Kriz, Jo{\~a}o Sedoc, Maria Kustikova, and Chris Callison-Burch. 2019.
\newblock \href {https://doi.org/10.18653/v1/P19-1365} {Comparison of diverse decoding methods from conditional language models}.
\newblock In \emph{Proceedings of the 57th Annual Meeting of the Association for Computational Linguistics}, pages 3752--3762, Florence, Italy. Association for Computational Linguistics.

\bibitem[{Lewis et~al.(2020)Lewis, Perez, Piktus, Petroni, Karpukhin, Goyal, K{\"u}ttler, Lewis, Yih, Rockt{\"a}schel et~al.}]{lewis2021retrievalaugmentedgenerationknowledgeintensivenlp}
Patrick Lewis, Ethan Perez, Aleksandra Piktus, Fabio Petroni, Vladimir Karpukhin, Naman Goyal, Heinrich K{\"u}ttler, Mike Lewis, Wen-tau Yih, Tim Rockt{\"a}schel, and 1 others. 2020.
\newblock Retrieval-augmented generation for knowledge-intensive nlp tasks.
\newblock \emph{Advances in Neural Information Processing Systems}, 33:9459--9474.

\bibitem[{Li et~al.(2023)Li, Zhang, Wang, Xiong, Lu, and Medioni}]{li2023gpt4recgenerativeframeworkpersonalized}
Jinming Li, Wentao Zhang, Tian Wang, Guanglei Xiong, Alan Lu, and Gerard Medioni. 2023.
\newblock \href {https://arxiv.org/abs/2304.03879} {Gpt4rec: A generative framework for personalized recommendation and user interests interpretation}.
\newblock \emph{Preprint}, arXiv:2304.03879.

\bibitem[{Lightman et~al.(2023)Lightman, Kosaraju, Burda, Edwards, Baker, Lee, Leike, Schulman, Sutskever, and Cobbe}]{lightman2023let}
Hunter Lightman, Vineet Kosaraju, Yuri Burda, Harrison Edwards, Bowen Baker, Teddy Lee, Jan Leike, John Schulman, Ilya Sutskever, and Karl Cobbe. 2023.
\newblock Let's verify step by step.
\newblock In \emph{The Twelfth International Conference on Learning Representations}.

\bibitem[{Lin(2004)}]{lin2004rouge}
Chin-Yew Lin. 2004.
\newblock \href {https://aclanthology.org/W04-1013/} {{ROUGE}: A package for automatic evaluation of summaries}.
\newblock In \emph{Text Summarization Branches Out}, pages 74--81, Barcelona, Spain. Association for Computational Linguistics.

\bibitem[{Massarelli et~al.(2020)Massarelli, Petroni, Piktus, Ott, Rockt{\"a}schel, Plachouras, Silvestri, and Riedel}]{massarelli2020decodingstrategiesaffectverifiability}
Luca Massarelli, Fabio Petroni, Aleksandra Piktus, Myle Ott, Tim Rockt{\"a}schel, Vassilis Plachouras, Fabrizio Silvestri, and Sebastian Riedel. 2020.
\newblock \href {https://doi.org/10.18653/v1/2020.findings-emnlp.22} {How decoding strategies affect the verifiability of generated text}.
\newblock In \emph{Findings of the Association for Computational Linguistics: EMNLP 2020}, pages 223--235, Online. Association for Computational Linguistics.

\bibitem[{Miao et~al.(2024)Miao, Oliaro, Zhang, Cheng, Wang, Zhang, Wong, Zhu, Yang, Shi, Shi, Chen, Arfeen, Abhyankar, and Jia}]{Miao_2024}
Xupeng Miao, Gabriele Oliaro, Zhihao Zhang, Xinhao Cheng, Zeyu Wang, Zhengxin Zhang, Rae Ying~Yee Wong, Alan Zhu, Lijie Yang, Xiaoxiang Shi, Chunan Shi, Zhuoming Chen, Daiyaan Arfeen, Reyna Abhyankar, and Zhihao Jia. 2024.
\newblock \href {https://doi.org/10.1145/3620666.3651335} {{SpecInfer}: Accelerating large language model serving with tree-based speculative inference and verification}.
\newblock In \emph{Proceedings of the 29th ACM International Conference on Architectural Support for Programming Languages and Operating Systems, Volume 3}, ASPLOS ’24. ACM.

\bibitem[{Nallapati et~al.(2016)Nallapati, Zhou, dos Santos, Gu{\ensuremath{\dot{}}}l{\c{c}}ehre, and Xiang}]{nallapati-etal-2016-abstractive}
Ramesh Nallapati, Bowen Zhou, Cicero dos Santos, {\c{C}}a{\u{g}}lar Gu{\ensuremath{\dot{}}}l{\c{c}}ehre, and Bing Xiang. 2016.
\newblock \href {https://doi.org/10.18653/v1/K16-1028} {Abstractive text summarization using sequence-to-sequence {RNN}s and beyond}.
\newblock In \emph{Proceedings of the 20th {SIGNLL} Conference on Computational Natural Language Learning}, pages 280--290, Berlin, Germany. Association for Computational Linguistics.

\bibitem[{Pham and Vo(2024)}]{pham2024reliablemedicalquestionanswering}
Duy~Khoa Pham and Bao~Quoc Vo. 2024.
\newblock \href {https://arxiv.org/abs/2408.13808} {Towards reliable medical question answering: Techniques and challenges in mitigating hallucinations in language models}.
\newblock \emph{Preprint}, arXiv:2408.13808.

\bibitem[{Qin et~al.(2025)Qin, He, Prakriya, Cong, and Sun}]{Qin_He_Prakriya_Cong_Sun_2025}
Zongyue Qin, Zifan He, Neha Prakriya, Jason Cong, and Yizhou Sun. 2025.
\newblock \href {https://doi.org/10.1609/aaai.v39i23.34690} {Dynamic-width speculative beam decoding for llm inference}.
\newblock \emph{Proceedings of the AAAI Conference on Artificial Intelligence}, 39(23):25056--25064.

\bibitem[{Spector and Re(2023)}]{spector2023accelerating}
Benjamin~Frederick Spector and Christopher Re. 2023.
\newblock \href {https://openreview.net/forum?id=RKHF3VYjLK} {Accelerating {LLM} inference with staged speculative decoding}.
\newblock In \emph{Workshop on Efficient Systems for Foundation Models @ ICML2023}.

\bibitem[{Vaswani et~al.(2017)Vaswani, Shazeer, Parmar, Uszkoreit, Jones, Gomez, Kaiser, and Polosukhin}]{NIPS2017_3f5ee243}
Ashish Vaswani, Noam Shazeer, Niki Parmar, Jakob Uszkoreit, Llion Jones, Aidan~N Gomez, \L~ukasz Kaiser, and Illia Polosukhin. 2017.
\newblock \href {https://proceedings.neurips.cc/paper_files/paper/2017/file/3f5ee243547dee91fbd053c1c4a845aa-Paper.pdf} {Attention is all you need}.
\newblock In \emph{Advances in Neural Information Processing Systems}, volume~30. Curran Associates, Inc.

\bibitem[{Wu et~al.(2016)Wu, Schuster, Chen, Le, Norouzi, Macherey, Krikun, Cao, Gao, Macherey, Klingner, Shah, Johnson, Liu, Łukasz Kaiser, Gouws, Kato, Kudo, Kazawa, Stevens, Kurian, Patil, Wang, Young, Smith, Riesa, Rudnick, Vinyals, Corrado, Hughes, and Dean}]{wu2016googlesneuralmachinetranslation}
Yonghui Wu, Mike Schuster, Zhifeng Chen, Quoc~V. Le, Mohammad Norouzi, Wolfgang Macherey, Maxim Krikun, Yuan Cao, Qin Gao, Klaus Macherey, Jeff Klingner, Apurva Shah, Melvin Johnson, Xiaobing Liu, Łukasz Kaiser, Stephan Gouws, Yoshikiyo Kato, Taku Kudo, Hideto Kazawa, and 12 others. 2016.
\newblock \href {https://arxiv.org/abs/1609.08144} {Google's neural machine translation system: Bridging the gap between human and machine translation}.
\newblock \emph{Preprint}, arXiv:1609.08144.

\end{thebibliography}
\clearpage

\appendix
\section{Empirical Analysis of Prefix Overlap}
\label{sec:prefix-overlap}
A key advantage of our trie-based beam search over conventional beam search is its ability to consolidate overlapping prefixes across beams, thereby reducing redundant KV cache storage. This naturally raises the question: how frequent is prefix overlap in practice, and are the observed efficiency gains justified?

In principle, certain prompts could yield little to no overlap, limiting the benefit of our method. However, extensive evaluations across multiple models, datasets, and tasks show that such cases are rare in practice.

To quantify overlap, we measured the ratio of memory usage between trie-based decoding ($M_T$) and conventional beam search ($M_B$), averaged across all samples ($\frac{M_T}{M_B}$).

Table~\ref{tab:prefix_overlap} summarizes results over 21 settings spanning three models from our main experiments (Llama 3.1 8B, Mistral-Small 24B, and Phi-3.5-mini 3.8B).
In addition, we report results for a much larger model, Llama 3.1 70B, on the HumanEval dataset with a beam width of 3. 
Due to computational limitations, this was the only feasible setting we could evaluate for the 70B model.

Across all settings, we observe an average ratio of 0.311 (median 0.274, standard deviation 0.155), consistently reflecting substantial memory savings. 
These findings indirectly confirm that significant prefix overlap is common in realistic scenarios, and that it underpins the efficiency gains of our approach. 
Importantly, at the 70B scale, trie-based decoding continued to deliver substantial savings, reinforcing its applicability to very large models.

\begin{table}[t]
\centering
\begin{tabular}{llrr}
\toprule
\textbf{Model} & \textbf{Dataset} & \textbf{$b$} & \textbf{$\frac{M_T}{M_B}$}\,$\downarrow$ \\
\midrule

\multirow{5}{*}{\textbf{Llama 3.1 8B}} & \multirow{2}{*}{CNN}        & 3  & 0.385 \\
             & & 9  & 0.158 \\
    \cmidrule{2-4}
             & \multirow{3}{*}{HumanEval}  & 3  & 0.669 \\
             & & 9  & 0.274 \\
             & & 15 & 0.212 \\
\midrule
\multirow{4}{*}{\textbf{Mistral-Small}} & \multirow{2}{*}{CNN}        & 3  & 0.398 \\
             & & 6  & 0.248 \\
            \cmidrule{2-4}
             & \multirow{2}{*}{HumanEval}  & 3  & 0.053 \\
             & & 6  & 0.065 \\
\midrule
\multirow{11}{*}{\textbf{Phi-3.5-mini}} & \multirow{2}{*}{CNN} & 3  & 0.386 \\
             & & 9  & 0.138 \\
            \cmidrule{2-4}
             & \multirow{3}{*}{HumanEval}  & 3  & 0.512 \\
             & & 9  & 0.302 \\
             & & 15 & 0.216 \\
             \cmidrule{2-4}
             & \multirow{3}{*}{GSM8K}      & 3  & 0.494 \\
             & & 9  & 0.274 \\
             & & 15 & 0.227 \\
             \cmidrule{2-4}
             & \multirow{3}{*}{WMT}        & 3  & 0.472 \\
             &  & 9  & 0.325 \\
             &  & 15 & 0.270 \\
            \midrule 
\textbf{Llama 3.1 70B} & HumanEval  & 3  & 0.455 \\
\bottomrule
\end{tabular}
\caption{Memory usage ratios of trie-based decoding relative to conventional beam search across models, datasets, and beam widths ($b$).}
\label{tab:prefix_overlap}
\end{table}

\section{Comparing Beam Search and Top-\texorpdfstring{$k$}{k} Sampling}

\begin{figure*}[b!]
    \centering

    \begin{lstlisting}[language=Python]
from typing import List

def parse_nested_parens(paren_string: str) -> List[int]:
    stack = []
    max_depths = []

    for char in paren_string:
        if char == '(':
            stack.append(len(stack))
        elif char == ')':
            if stack:
                max_depth = stack.pop()
                max_depths.append(max_depth + 1)
                
    return max_depths
\end{lstlisting}

\begin{lstlisting}[language=Python]
from typing import List

def parse_nested_parens(paren_string: str) -> List[int]:
    result = []
    for group in paren_string.split():
        depth = 0
        max_depth = 0
        for char in group:
            if char == '(':
                depth += 1
                max_depth = max(max_depth, depth)
            elif char == ')':
                depth -= 1
        result.append(max_depth)
    return result
    \end{lstlisting}
    \caption{Code generated with top-$k$ sampling (upper) vs. beam search (lower)}
    \label{fig:python-samples}
\end{figure*}

While our primary objective is to improve the efficiency of beam search decoding, it is also informative to compare beam search with alternative strategies to better understand its relative strengths and limitations.

We conducted additional experiments on the HumanEval benchmark using the Phi-3.5-mini model, comparing beam search with top-$k$ sampling ($k=50$). Top-$k$ sampling achieved an accuracy of 64\%, which is lower than greedy decoding (65\%) and beam search (70\% with $b=15$). 
Although top-$k$ sampling matches greedy decoding in speed, its stochastic nature introduces greater variability and increases susceptibility to errors. 

As an illustration, consider generating a Python function to compute the maximum depth of nested parentheses for each space-separated substring (e.g., input ``\texttt{(()()) ((())) ()}'' $\rightarrow$ output \texttt{[2, 3, 1]}). 
As shown in Figure~\ref{fig:python-samples}, with top-$k$ sampling, the generated code erroneously measures the depth of each individual parenthesis pair, producing \texttt{[2, 2, 1, 3, 2, 1, 1]}. 
In contrast, the code produced via beam search correctly aggregates per substring and returns \texttt{[2, 3, 1]}.

\end{document}